\documentclass[a4paper,twoside]{article}

\usepackage{epsfig}
\usepackage{subcaption}
\usepackage{calc}
\usepackage{amssymb}
\usepackage{amstext}
\usepackage{amsmath}
\usepackage{amsthm}
\usepackage{multicol}
\usepackage{pslatex}
\usepackage{apalike}
\usepackage{algorithm2e}
\usepackage[bottom]{footmisc}
\usepackage{graphicx}
\usepackage{xcolor}
\usepackage{booktabs}
\usepackage[inline]{enumitem}
\usepackage{url}
\usepackage{listings}

\usepackage{SCITEPRESS}     

\lstdefinelanguage{yaml}{
  keywords={true,false,null,y,n},
  keywordstyle=\color{blue}\bfseries,
  basicstyle=\ttfamily\small,
  sensitive=false,
  comment=[l]{\#},
  commentstyle=\color{green!50!black},
  morestring=[b]",
  morestring=[b]',
  stringstyle=\color{red},
}

\newcommand{\para}[1]{\vspace{5pt} \noindent \textbf{#1}}

\usepackage{lipsum}

\newcommand\blfootnote[1]{%
  \begingroup
  \renewcommand\thefootnote{}\footnote{#1}%
  \addtocounter{footnote}{-1}%
  \endgroup
}

\begin{document}
\title{COntExt: Towards Context-Aware Ontology Extension from Operational Metrics}

\author{\authorname{Hussain Hussain\sup{1}\orcidAuthor{0000-0002-0959-623X}, Stefan Schöberl\sup{2}\orcidAuthor{0009-0006-0245-3558}, Angelika Schneider\sup{3}\orcidAuthor{0000-0002-8962-3276} and Verena Geist\sup{2}\orcidAuthor{0000-0002-3729-1265}}
\affiliation{\sup{1}Know Center Research GmbH, Graz, Austria}
\affiliation{\sup{2}Software Competence Center Hagenberg GmbH, Hagenberg im Mühlkreis, Austria}
\affiliation{\sup{3}Fraunhofer AISEC, Garching bei München, Germany}
\email{hhussain@know-center.at, stefan.schoeberl@scch.at, \\angelika.schneider@aisec.fraunhofer.de, verena.geist@scch.at}
}


\abstract{
Organizations increasingly define operational metrics  in structured, machine-readable formats to monitor systems, processes, and compliance. These metric definitions implicitly encode domain knowledge, such as referencing concepts, properties, and relationships, that often extends what is captured in formal ontologies. Yet the connection between operational metric catalogues and ontological knowledge remains manual, ad-hoc, and labour-intensive. We present COntExt, a framework for context-aware ontology extension that takes structured metric definitions as input and suggests how referenced concepts and properties should be integrated into an existing ontology, utilizing the context of these metrics. The framework defines the extension problem as three sub-tasks:  parent class prediction, relation type prediction, and data property assignment. Across seven ontologies spanning four domains, we evaluate different algorithms for each task. Our results show that metric-derived context improves the suggestions over ontology-context baselines for relation type prediction and data property assignment. Our work demonstrates that operational metric catalogues are a practical and underexploited source for ontology extension.
This work enables organizations to maintain their ontologies at a significantly lower cost than manual engineering.
}

\onecolumn \maketitle \normalsize \setcounter{footnote}{0} \vfill

\section{\uppercase{Introduction}}
\label{sec:introduction}
\para{Motivation.}\blfootnote{\textit{Preprint; a short version is accepted at KEOD 2026.}}
Domain ontologies provide formal, shared conceptualizations that enable interoperability, reasoning, and knowledge reuse across systems.
In practice, however, ontologies lag behind the domains they represent.
New concepts, properties, and relationships emerge continuously, driven by evolving regulations, operational requirements, and technological change~\cite{zablith2015ontology,elnagar_automatic_nodate}.
Ontology engineers must manually identify these gaps and extend the ontology, a process that demands both domain expertise and knowledge-engineering skill.
Meanwhile, many organizations already capture significant domain knowledge in a parallel, less formal artefact: structured metric definitions~\cite{van2016business,yu2026ontometricontologydrivenllmassistedframework}.
Metric definitions, expressed in semi-structured human-readable forms, represent an important intermediate artefact between natural-language text and formal ontologies.
A cybersecurity team, for example, may maintain hundreds of YAML-based metric definitions that reference threat types, system components, compliance properties, and their relationships, often using terminology that closely mirrors the organization's domain ontology.
However, this implicit knowledge is currently discarded from the perspective of ontology maintenance.

\begin{figure*}[t]
    \centering
    \includegraphics[width=\textwidth]{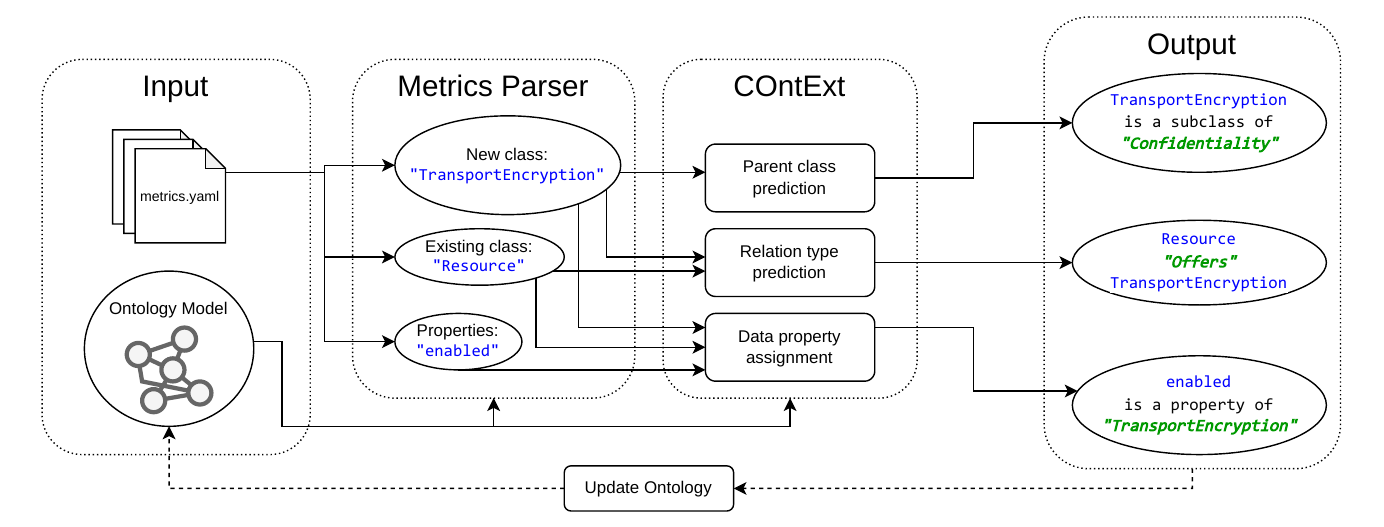}
    \caption{An illustration of COntExt framework. This figure shows the \textit{input} to COntExt, represented by the ontology model and the operational metric files in \texttt{.yaml} format. The \textit{metrics parser} identifies mentions of concepts (classes and properties). The identified concepts are then passed on to \textit{COntExt}, which performs the corresponding tasks: 1) predicting the parent of an identified class, 2) predicting the type of relation between a pair of classes, and 3) predicting the class that owns the mentioned property. The \textit{output} of COntExt can be passed on to update the ontology. The example metric (shown in Listing~\ref{listing:metric}) is extracted from security criteria “Encryption of data for transmission”,  specified in the BSI C5:2020 (CRY-02).}
    \label{fig:framework}
\end{figure*}

\para{Problem.}
Existing approaches to ontology completion draw on two main sources: unstructured text corpora \cite{liu_concept_2020,sanagavarapu2022ontoenricher,kollapally2025ontology} and the structural patterns within the ontology itself \cite{shen_taxoexpan_2020,jiang2022taxoenrich,ma2021hyperexpan}.
Neither exploits the middle ground occupied by operational artefacts such as metric definitions.
These artefacts are more constrained and semantically explicit than free text, yet they are richer in contextual information than bare ontology axioms.

\para{Approach.}
Our approach centres on a key question: \emph{Does metric-derived context (e.g., descriptions, comments, and configuration values embedded in structured metric catalogues) provide ontology-extension signals beyond an ontology's own linguistic and structural content?}
We address this question by building a \textbf{C}ontext-aware \textbf{Ont}ology \textbf{Ext}ension framework (COntExt) that tackles three extension sub-tasks: 1) parent class prediction, 2) relation type prediction, and 3) data property assignment.
COntExt is an algorithm-agnostic framework that allows for injecting textual context with each query, making it suitable for metric-driven context.
Figure~\ref{fig:framework} illustrates COntExt framework and its sub-tasks with an example of a metric-driven query.
We evaluate the utility of extension methods for each task using only ontology context and then comparing it to layering metric-derived context on top of the same methods.
We demonstrate the benefit of metrics-derived context for ontology extension on an ontology associated with a metric catalogue, \emph{CertGraph ontology}~\cite{schoeberl2024certgraph}.
Our findings suggest that operational metrics are useful sources of context for relation type prediction and data property assignment on the evaluated use case.
However, the context seems to degrade the extension accuracy for parent class prediction.
Our source code is publicly available\footnote{\url{https://github.com/sqrhussain/COntExt}} and can be easily adapted for similar use cases. 

\para{Contribution.}
We make the following contributions:
\begin{enumerate}
    \item  A metric-driven algorithm-agnostic ontology extension framework (COntExt) that takes the target ontology and metric definitions as input and suggests their integration into an existing ontology. 
    \item A context enrichment strategy that incorporates metric descriptions, comments, configuration values, and ontology annotations into algorithm inputs, improving suggestion relevance beyond ontology-only baselines.
    \item  A comparative evaluation of several algorithms across the three sub-tasks and seven ontologies.
    \item  Empirical evidence that structured operational metric catalogues are a viable source for ontology extension, bridging the gap between operational knowledge and formal representations.
\end{enumerate}

\para{Impact.}
Our work exploits metric definitions for ontology extension, opening an opportunity for exploring a class of input, typically overlooked by existing literature.
With COntExt, organizations maintaining both a domain ontology and a structured metric catalogue can keep their ontology aligned with evolving operational definitions at substantially lower cost than manual ontology engineering.

\section{Related Work}

\subsection{Ontology Extension, Completion, and Enrichment}
The interest in maintaining domain ontologies to keep them up to date has been growing for years.
A large body of work focuses on adding new classes or relations based on textual corpora.
\cite{elnagar_automatic_nodate} develop a domain-independent framework that generates formal ontologies from unstructured text corpus.
\cite{sanagavarapu2022ontoenricher} use deep sequential models, such as Bi-LSTMs, to extract concepts and relations from unstructured text and enrich a base security ontology.
Several works focus on language models at different stages of ontology enrichment.
For example, \cite{kollapally2025ontology} use large language models (LLMs) for extracting triples from textual context.
\cite{dong_language_2024} evaluate pretrained language models (PLMs) embeddings and LLMs for relation matching and prediction.
\cite{liu_concept_2020} use BERT next-sentence prediction on ontology neighbourhoods to predict parents of biomedical concepts.
Finally, KROMA \cite{nguyen2025kroma} integrates LLMs with a knowledge retrieval pipeline for concept semantic equivalence.
These examples show that contextual language models can capture ontological semantics and support automated insertion of novel terms.

Another set of relevant works focuses on graph-based completion methods.
\cite{chen2025ontology} provide a comprehensive survey of ontology embedding methods.
Several graph neural networks (GNN)-based approaches focus on predicting the parent concept.
TaxoExpan \cite{shen_taxoexpan_2020} and TaxoEnrich \cite{jiang2022taxoenrich} learn from existing hierarchies via self-supervised objectives: they use position-aware GNNs to predict where new hyponyms should attach.
HyperExpan \cite{ma2021hyperexpan} similarly leverages hyperbolic embeddings to model hierarchical structure. 
For predicting missing relations, \cite{meznar_ontology_2022} uses structure-only link-prediction algorithms. 
With the exception of \cite{meznar_ontology_2022}, all these methods combine \emph{structural features} (position in a hierarchy) with \emph{semantic features of surface text}, e.g., word meanings, to rank candidates.

Our proposed framework incorporates these ontology extension tasks previously tackled by these works.
However, instead of unstructured text corpora or pure graph structure with surface text, our framework enriches predictions with features from operational metric definitions that represent the most recent domain knowledge.

\subsection{Ontology and Metrics}
Research on ontology and metrics has proliferated across domains such as semantic web, IoT, biomedical, engineering, and sustainability.

On the one hand, a line of research uses ontologies to define or construct metrics. This line of work produced several 
ontology-based frameworks for cohesive measurement of health and quality of life \cite{cella2022patient},
harmonizing measures for bike (transport) network evaluations \cite{grisiute2024ontology},
evaluating security assurance metrics \cite{wen2024ontology}, and 
guiding extraction of ESG metrics \cite{yu2026ontometricontologydrivenllmassistedframework}.

On the other hand, a more related line of work uses metrics for assessing ontologies (or related schemas).
These works treat \emph{ontology metrics} as formal, quantitative measures of ontology structure or content.
A recurring trend is the creation of tools and frameworks for standardization.
For example, NEOntometrics \cite{reiz2024neontometrics} implement dozens of proposed metrics, addressing a historical gap of “missing implementations”, while OntoInsight \cite{sammi2025ontoinsight} guides quality evaluation with metrics and AI-driven recommendations.

Others tackle domain-specific needs, e.g., a cybersecurity ontology \cite{bryniarska2022ontology} uses standard metrics to assess security models, MEDTO \cite{hao_medto_2021} maps concepts from medical databases, used by medical experts, to an ontology to support downstream applications, a survey by \cite{bain2024systematic} evaluates 24 COVID-19 ontologies via structural assessment, OntoLogX \cite{cotti2026ontologx} transforms raw cyber threat intelligence logs into ontology-grounded knowledge graphs, and SHACLEval \cite{reiz2025shacleval} defines metrics specific to SHACL constraints to ensure data graph quality.
More specifically, the cybersecurity community has produced extensive catalogues and classifications of security metrics, proposing and refining taxonomies for security metrics \cite{savola2007towards,pendleton2016survey,longueira2020quantify,morrison2018mapping}.
Furthermore, metric-driven ontology engineering has also proliferated in the context of continuous certification and measurement \cite{stephanow2015towards,schoeberl2024certgraph}.

Despite the increasing adoption of operational metrics and their association with domain ontologies, previous work has not yet investigated leveraging operational metrics to streamline or automate ontology development.
Although some works, such as the one from \cite{sammi2025ontoinsight}, have used metrics to guide ontology development, these works focus on refining the quality of the ontology and not its alignment with developing domain knowledge.
Our work tackles this gap in particular by developing a framework that allows exploiting operational metrics as context for the extension of domain ontologies to keep ontologies up to date with the domain they represent.

\section{Framework}
\label{sec:framework}

This section describes the COntExt framework at the conceptual level: what it takes as input, how it decomposes the ontology extension problem, and how the context enrichment is applied.
The specific algorithms are detailed in the evaluation (Section~\ref{sec:evaluation}).

\subsection{Input Representation}
\para{Ontology model.} COntExt operates on any formal ontology that defines a class hierarchy, object properties (relationships between classes), and data properties (attributes of classes). The model of the input ontology consists of:

\begin{itemize}
  \item A class hierarchy with parent-child relationships.
  \item An inventory of object properties and data properties.
  \item Property axioms specifying domain and range restrictions. These axioms link classes to each other (through object property axioms) and classes to attributes (through data property axioms).
\end{itemize}

The ontology may be supplied in any machine- and human-readable format (e.g., OWL/XML, RDF, OWX) from which these structural elements can be extracted.
Textual context can be also harnessed from ontologies if available, e.g., in the form of \texttt{rdfs:comment}.
Studying methods for extraction and model building is not within the scope of our work.

\para{Metric definitions.} The second input is one or more structured metric definitions. Each definition is expected to be in a machine-readable format (e.g., YAML, JSON, XML) and to contain fields that reference domain concepts, either explicitly (e.g., through typed references to classes and properties) or implicitly (e.g., through descriptions, configuration parameters, and comments that mention domain terminology). COntExt requires a parser that can extract candidate class references and property references from these definitions.
The specific extraction mechanism is not the topic of our work, but it depends on the format and conventions of the metric catalogue in use.





\subsection{Extension Tasks}

COntExt decomposes the ontology extension problem into three independent tasks.
Each task takes a specific type of input (or "query"), applies a pluggable suggestion algorithm, and produces a ranked list of candidates.
Let us assume a set of classes $\mathcal{C}$, a set of relation types $\mathcal{R}$, and a set of data properties (attributes) $\mathcal{A}$ that may be associated with classes.
The ontology is composed of 1) a class hierarchy $(c, \text{is-a}, p) \in \mathcal{C} \times \mathcal{C}$, 2) triples of relations $(c_i,r,c_j) \in \mathcal{C} \times \mathcal{R}\times \mathcal{C}$, and 3) pairs of attribute associations $(c, a) \in \mathcal{C} \times \mathcal{A}$.

\para{Task 1 --- Parent class prediction.}
Given a query of a class name $c \notin \mathcal{C}$, rank all existing classes $p \in \mathcal{C}$ by their suitability as parent $(c, \text{is-a}, p)$ in the hierarchy and present the top-$k$ suggestions.
Any algorithm that can score the semantic fit of a subclass relation between a candidate term and existing ontology classes can serve as the suggestion engine.

\para{Task 2 --- Relation type prediction.} Given a query of two classes $c_i, c_j \in \mathcal{C}$, rank all candidate relation types $r \in \mathcal{R}$ by the plausibility of the triple $(c_i, r, c_j)$. The framework evaluates both directions, $(c_i, r, c_j)$ and $(c_j, r, c_i)$, and returns a ranked list of the directed relation types. Any algorithm that can score the plausibility of a (subject, predicate, object) triple can be plugged into this task.

\para{Task 3 --- Data property assignment.} Given a query of a data property $a \in \mathcal{A}$ with inferred data type $\tau$ and a set of candidate owner classes $\{c_1, \ldots, c_m\} \subset \mathcal{C}$, predict which class $c^*$ from the candidates should own data property $a$, i.e., the plausibility of the association pair $(c^*,a)$. The data type $\tau$ may be inferred from the metric's configuration block if this context was available, e.g., boolean from true/false target values, numeric from thresholds. Any algorithm capable of scoring the association between a property and a set of candidate classes can be used.

\subsection{Context Enrichment}
Depending on the richness of the input data, the framework can optionally enrich suggestion inputs with context beyond bare term names.
We distinguish between two sources of additional context:

\begin{itemize}
  \item \textbf{Ontology-derived context:} Annotations (e.g., \texttt{rdfs:comment}, \texttt{rdfs:label}) attached to existing ontology classes and properties, which provide definitional context.
  \item \textbf{Metric-derived context:} Descriptions, comments, configuration values, and other fields from the metric definition that provide semantic grounding for the referenced concepts.
\end{itemize}

Context enrichment is optional and configurable per task, allowing systematic comparison of how much additional signal operational artefacts provide beyond ontology structure alone.



\section{Evaluation}
\label{sec:evaluation}

This section presents the evaluation of COntExt. We first state the research questions that guide the evaluation (Section~\ref{sec:rqs}), then describe the ontologies and metric corpus used for evaluation (Sections~\ref{sec:ontologies}--\ref{sec:metric-corpus}), define the evaluation protocols for each extension task (Section~\ref{sec:protocols}), introduce the algorithms used for each task (Section~\ref{sec:algorithms}) and report results organized by research question (Section~\ref{sec:results}).

\begin{table*}[t]
\small
\centering
\caption{Ontologies used in the evaluation.}
\label{tab:ontologies}
\begin{tabular}{llrrrrrr}
\toprule
\textbf{Name}     &    \textbf{Domain}              &\textbf{Classes}&  \textbf{Relations}&  \textbf{Relation Types}&  \textbf{Classes with Attributes} &  \textbf{Hierarchy Depth} \\
\midrule
Pizza & Educational        & 86 & 170 & 3 & 0 &  6 \\
FIBO$_{BE}$ & Finance            & 248 & 87 & 43 & 3 &  6 \\
SAREF$_{ener}$ & IoT (Energy)       & 90 & 117 & 87 & 31  & 4 \\
JRC & Cybersecurity      & 204 & 1350 & 9 & 7 &  2 \\
CSO$_{sec}$ & Cybersecurity      & 1055 & 2544 & 2 & 0  & 8 \\
TAC & Cybersecurity      & 176 & 94 & 62 & 382  & 6 \\
CertGraph    & Cybersecurity      & 264 & 161 & 20 & 104  & 6 \\
\bottomrule
\end{tabular}
\end{table*}

\subsection{Research Questions}
\label{sec:rqs}

\begin{description}
  \item[RQ1 --- Algorithm comparison.] How do different algorithms compare across the three extension tasks? This question evaluates multiple algorithms for each task on all ontologies.
  \item[RQ2 --- Context enrichment.] Does enriching algorithm inputs with ontology annotations and metric-derived context (descriptions, comments, configuration values) improve suggestion quality over baselines that only use term names as linguistic features? How does the metric-derived context compare to the ontology context?
\end{description}
These questions are evaluated on all three tasks using the applicable datasets and algorithms for each.

\subsection{Ontologies}
\label{sec:ontologies}

We evaluate the framework on seven ontologies that vary in size, annotation richness, and domain, with a focus on the cybersecurity domain.
Table~\ref{tab:ontologies} summarizes their key characteristics.

Pizza\footnote{\url{https://protege.stanford.edu/ontologies/pizza/pizza.owl}}
is a small ontology intended for educational purposes about OWL/RDF ontologies.
FIBO\footnote{\url{https://edmconnect.edmcouncil.org/okgspecialinterestgroup/resources-sig-link/resources-sig-link-fibo-products-download}} (Financial Industry Business Ontology)~\cite{Bennett2013FIBO} is a formal ontology that provides a common vocabulary for financial contracts and related concepts, comprising several sub-ontologies.
We use the sub-ontology for Business Entities (BE) domain FIBO$_{BE}$, which defines business concepts that are used for data governance, interoperability, and in regulatory reporting about business entities.
The Smart Applications REFerence (SAREF)\footnote{\url{https://saref.etsi.org/}}~\cite{garcia2023etsi} is intended for interoperability between different components among various sectors in the Internet of Things (IoT). We take the subset ontology SAREF$_{ener}$, which is the extension for the Energy domain.
JRC\footnote{\url{https://cybersecurity-atlas.ec.europa.eu/cybersecurity-taxonomy}}
is a cybersecurity taxonomy from the European Commission’s Joint Research Centre.
CSO\footnote{\url{https://cso.kmi.open.ac.uk/home}}~\cite{Salatino2019CSOClassifier} is a large-scale ontology of research areas that was automatically generated from the Rexplore dataset, which consists of about 16 million publications.
Threat Actor Context (TAC)\footnote{\url{https://github.com/oasis-tcs/tac-ontology}}
ontology is a semantic representation of cyber threat intelligence built on top of the STIX 2.1 standard to enable interoperability across different threat-intelligence-relevant information sources and technological solutions.
Finally, CertGraph\footnote{\url{https://github.com/Cybersecurity-Certification-Hub/security-metrics}}~\cite{schoeberl2024certgraph} enables the automated assessment of certification-relevant security metrics by providing a semantic representation of security concepts, relationships, and evidence sources.
CertGraph is also associated with a catalogue of security metrics, offering a metric-rich use case for the evaluation of metric-derived ontology extension.

In our reference implementation, ontologies are parsed directly from OWL/XML with the \texttt{lxml} library.
We also implement light per-source converters (under \texttt{data\_transformation/}) normalising other formats such as Pizza, SAREF, FIBO, and CSO into the same OWX representation before parsing.

\begin{lstlisting}[
    language=yaml,
    caption={Metric example from CertGraph security metrics.},
    label={listing:metric}
]
name: TransportEncryptionEnabled
description: This rule assesses whether a [Resource] has [TransportEncryption] [p1:enabled] correctly configured.
category: TransportEncryption
version: "v1" 
comments: Transport encryption is a standard practice to keep sensitive data confidential when it is transmitted.
configuration:
  p1:
    operator: "=="
    targetValue: True
\end{lstlisting}

\subsection{Metric Corpus}
\label{sec:metric-corpus}
The metric corpus consists of structured cybersecurity metric definitions associated with the \emph{CertGraph} ontology. The corpus spans 12 security categories (e.g., AI Security, Identity Management, Transport Encryption) and contains 64 individual metric definitions in YAML format. Each metric references domain concepts through its description, configuration block, and optional comments.
This corpus is used exclusively for RQ2 to test whether metric-derived context improves suggestion quality.
References are tagged inline within the \texttt{description} field --- bare \texttt{[Term]} for classes and \texttt{[pN:propertyName]} for properties --- and extracted with a single regular expression.
Listing~\ref{listing:metric} shows an example for a metric associated with CertGraph.


\subsection{Evaluation Protocols}
\label{sec:protocols}
Each extension task is evaluated using a leave-one-out protocol that masks known ontology structure and measures whether the framework can recover it.
This approach avoids the need for manually labelled ground truth and allows evaluation on any ontology.
For RQ2, only the concepts, relations, and properties that are mentioned in the metric files are considered for evaluation.

\para{Parent class prediction (Task 1).}
For each class $c$ in the ontology with a known parent $p$, we temporarily remove $c$ from the hierarchy and ask the suggestion engine to rank all remaining classes by their suitability as parent of $c$. We report:

\begin{itemize}
  \item \textbf{Mean Reciprocal Rank (MRR):} the average of $1/\text{rank}$ of the correct parent across all test cases.
  \item \textbf{Hits@$k$} ($k \in \{1, 3, 5, 10\}$): the proportion of test cases where the correct parent appears in the top $k$ suggestions.
\end{itemize}

\para{Relation type prediction (Task 2).}
For each object relation axiom $(d, r, g)$ in the ontology, where $d$ is the domain class, $r$ is the relation type, and $g$ is the range class, we remove the axiom and ask the suggestion engine to rank all candidate relation types for the query pair $(d, g)$.
We report MRR and Hits@$k$ as above, with the candidate set being the full inventory of relation types in the ontology.

\para{Data property assignment (Task 3).}
For each data property axiom $(c, a)$ --- where class $c$ owns data property (attribute) $a$ --- we construct a binary classification task. We select a \emph{neighbour} class $c'$ that is structurally related to $c$ (via shared object properties or adjacency in the hierarchy) but does not own $a$. The suggestion engine must identify the correct owner from the candidate set $\{c, c'\}$. We report \emph{accuracy} over all non-trivial cases (i.e., those where a valid neighbour distractor exists).
The neighbour-based distractor selection is deliberately harder than random-negative sampling: because $c'$ is structurally close to $c$, the task requires the algorithm to distinguish between conceptually related classes rather than trivially dissimilar ones.

\para{Significance testing.}
\label{sec:significance}
To separate genuine differences between algorithms from sampling and training noise, we compare any two algorithms on the same task with a test paired at the level of individual masked cases: a two-sided approximate-randomization (permutation) test, Holm--Bonferroni-corrected across the ontologies in the comparison, together with a per-seed sign-consistency check that exposes training (seed) variance.
The per-case score is the reciprocal rank for the ranking tasks (parent class prediction and relation type prediction) and the accuracy for data property assignment.
We label a difference \emph{robust} only when the Holm-adjusted p-value is below $0.05$ and the sign of the per-case mean difference is consistent across every training seed; otherwise the gap is reported as a statistical tie or as within training noise.

\subsection{Algorithms and Configurations}
\label{sec:algorithms}

For each task, we evaluate a set of supervised state-of-the-art methods.
To evaluate context enrichment, we also implement unsupervised heuristics on top of pre-trained language models, such as BERT and SentenceTransformer, that have the capacity of including additional context.
The context is the \emph{prose text} existing either in the ontology annotations associated with the class, e.g.,  \texttt{rdfs:comment}, or in the textual description of the metric file corresponding to the query.
We evaluate all algorithms for RQ1 (algorithm comparison), while for RQ2 (context enrichment), we only evaluate heuristics that support context enrichment.
The adaptation of existing supervised methods is beyond the scope of this current work.

\subsubsection{Task 1 --- Parent class prediction}
\para{BERT Fine-tuning~\cite{liu_concept_2020}.} A BERT model is fine-tuned as a binary classifier on (child, parent) concept pairs from the ontology hierarchy. Positive pairs are real parent-child relationships; negatives are structurally similar non-parent nodes. At inference, all candidates are scored by the classifier's IS-A probability. We use this method as an example for language-model-based parent class prediction -- \textit{no context enrichment is adapted for this method}.

\para{TaxoExpan~\cite{shen_taxoexpan_2020}.} A position-enhanced Graph Attention Network encodes local ego-graphs around each candidate parent position. Each ego-graph contains the candidate node, its parent, and its children, with learned position embeddings distinguishing these roles. A log-bilinear matching model scores the compatibility between the ego-graph encoding and the query concept embedding. The model is trained with InfoNCE contrastive loss. This method represents an example of GNN-based parent class prediction -- \textit{no context enrichment}.


\para{Child aggregation of SentenceTransformer embedding (ChildAgg).}
All ontology class names are embedded into dense vectors using a sentence-transformer~\cite{reimers2019sentencebert} model -- we use \texttt{all-MiniLM-L6-v2} model.
For each class, we replace its embedding with the mean of its own and its children's embeddings.
For a candidate class, the framework computes cosine similarity against all existing class embeddings and returns the top-$k$ most similar classes as parent suggestions.
We develop this heuristic to be able to enrich it with linguistic context extracted from metrics and ontology annotations.
To enrich this method, we append the textual context to the class name before encoding.

\begin{table*}[!t]
\small
\centering
\caption{RQ1 --- Parent class prediction: algorithm comparison across ontologies. Mean$\pm$std over 5 random seeds. Best score per ontology is in boldface, and statistically significant wins are underlined.}
\label{tab:rq1-parent}
\begin{tabular}{@{}lllllll@{}}
\toprule
\textbf{Ontology} & \textbf{Algorithm}    & \textbf{MRR}        & \textbf{Hits@1}     & \textbf{Hits@3}     & \textbf{Hits@5}     & \textbf{Hits@10}    \\ \midrule
Pizza             & BERT      & $0.38_{\pm 0.09}$            & $0.28_{\pm 0.06}$            & $0.41_{\pm 0.15}$            & $0.49_{\pm 0.17}$            & $0.67_{\pm 0.14}$            \\
Pizza             & TaxoExpan       & $0.56_{\pm 0.06}$            & $0.47_{\pm 0.06}$            & $0.64_{\pm 0.07}$            & $0.68_{\pm 0.08}$            & $0.76_{\pm 0.08}$            \\
Pizza             & ChildAgg         & $\mathbf{0.68_{\pm 0.07}}$   & $\mathbf{0.51_{\pm 0.09}}$   & $\mathbf{0.82_{\pm 0.06}}$   & $\mathbf{0.94_{\pm 0.05}}$   & $\mathbf{0.96_{\pm 0.05}}$   \\\midrule
FIBO$_{BE}$          & BERT      & $0.04_{\pm 0.02}$            & $0.01_{\pm 0.02}$            & $0.04_{\pm 0.02}$            & $0.09_{\pm 0.04}$            & $0.15_{\pm 0.06}$            \\
FIBO$_{BE}$          & TaxoExpan       & $0.20_{\pm 0.04}$            & $0.10_{\pm 0.01}$            & $0.25_{\pm 0.06}$            & $0.33_{\pm 0.07}$            & $0.44_{\pm 0.07}$            \\
FIBO$_{BE}$          & ChildAgg         & $\mathbf{0.26_{\pm 0.03}}$   & $\mathbf{0.12_{\pm 0.03}}$   & $\mathbf{0.34_{\pm 0.07}}$   & $\mathbf{0.46_{\pm 0.08}}$   & $\mathbf{0.63_{\pm 0.05}}$   \\\midrule
SAREF$_{ener}$       & BERT      & $0.64_{\pm 0.11}$            & $\mathbf{0.60_{\pm 0.12}}$   & $0.64_{\pm 0.10}$            & $0.70_{\pm 0.12}$            & $0.76_{\pm 0.14}$            \\
SAREF$_{ener}$       & TaxoExpan       & $\mathbf{0.66_{\pm 0.09}}$   & $0.53_{\pm 0.15}$            & $\mathbf{0.76_{\pm 0.06}}$   & $\mathbf{0.82_{\pm 0.06}}$   & $0.87_{\pm 0.06}$            \\
SAREF$_{ener}$       & ChildAgg         & $0.48_{\pm 0.11}$            & $0.31_{\pm 0.13}$            & $0.57_{\pm 0.11}$            & $0.72_{\pm 0.08}$            & $\mathbf{0.93_{\pm 0.06}}$   \\\midrule
JRC        & BERT      & $0.17_{\pm 0.01}$            & $0.06_{\pm 0.03}$            & $0.24_{\pm 0.07}$            & $0.34_{\pm 0.06}$            & $0.44_{\pm 0.08}$            \\
JRC        & TaxoExpan       & $0.32_{\pm 0.03}$            & $0.22_{\pm 0.07}$            & $0.40_{\pm 0.04}$            & $0.44_{\pm 0.04}$            & $0.47_{\pm 0.06}$            \\
JRC        & ChildAgg         & \underline{$\mathbf{0.47_{\pm 0.03}}$}   & $\mathbf{0.29_{\pm 0.06}}$   & $\mathbf{0.59_{\pm 0.05}}$   & $\mathbf{0.72_{\pm 0.03}}$   & $\mathbf{0.85_{\pm 0.04}}$   \\\midrule
CSO$_{sec}$          & BERT      & $0.06_{\pm 0.03}$            & $0.03_{\pm 0.02}$            & $0.06_{\pm 0.03}$            & $0.08_{\pm 0.04}$            & $0.15_{\pm 0.06}$            \\
CSO$_{sec}$          & TaxoExpan       & \underline{$\mathbf{0.32_{\pm 0.02}}$}   & $\mathbf{0.18_{\pm 0.03}}$   & $\mathbf{0.41_{\pm 0.01}}$   & $\mathbf{0.50_{\pm 0.01}}$   & $\mathbf{0.63_{\pm 0.03}}$   \\
CSO$_{sec}$          & ChildAgg         & $0.20_{\pm 0.03}$            & $0.09_{\pm 0.02}$            & $0.24_{\pm 0.04}$            & $0.36_{\pm 0.03}$            & $0.52_{\pm 0.02}$            \\\midrule
TAC               & BERT      & $0.18_{\pm 0.05}$            & $0.07_{\pm 0.05}$            & $0.22_{\pm 0.14}$            & $0.29_{\pm 0.12}$            & $0.49_{\pm 0.05}$            \\
TAC               & TaxoExpan       & $0.44_{\pm 0.09}$            & $0.31_{\pm 0.10}$            & $0.55_{\pm 0.06}$            & $0.66_{\pm 0.08}$            & $0.70_{\pm 0.08}$            \\
TAC               & ChildAgg         & $\mathbf{0.52_{\pm 0.05}}$   & $\mathbf{0.35_{\pm 0.08}}$   & $\mathbf{0.63_{\pm 0.10}}$   & $\mathbf{0.74_{\pm 0.06}}$   & $\mathbf{0.86_{\pm 0.02}}$   \\\midrule
CertGraph         & BERT      & $0.12_{\pm 0.04}$            & $0.08_{\pm 0.03}$            & $0.14_{\pm 0.06}$            & $0.16_{\pm 0.04}$            & $0.24_{\pm 0.07}$            \\
CertGraph         & TaxoExpan       & $0.32_{\pm 0.06}$            & $0.18_{\pm 0.05}$            & $0.40_{\pm 0.08}$            & $0.48_{\pm 0.06}$            & $0.64_{\pm 0.04}$            \\
CertGraph         & ChildAgg         & $\mathbf{0.38_{\pm 0.03}}$   & $\mathbf{0.24_{\pm 0.02}}$   & $\mathbf{0.45_{\pm 0.06}}$   & $\mathbf{0.61_{\pm 0.05}}$   & $\mathbf{0.72_{\pm 0.07}}$   \\ \bottomrule
\end{tabular}%

\end{table*}

\subsubsection{Task 2 --- Relation type prediction}

\para{BertConvE~\cite{liu2022effective}.} A small BERT model is trained from scratch on random walks extracted from the ontology's graph structure. Each entity and relation is a single token in a custom vocabulary. At inference, the relation position in ``$c_i$ [MASK] $c_j$'' is predicted via masked language modelling. BertConvE is an advanced knwoledge graph completion method that uses masked language model (MLM)-like training, but it is merely structural as the nodes are thought of as tokens. \textit{Context enrichment is not applicable here.}

\para{TransE~\cite{bordes2013translating}.} A translating-embedding model learns entity and relation vectors from the ontology's existing triples such that $\mathbf{h} + \mathbf{r} \approx \mathbf{t}$ for valid triples $(h, r, t)$. For a class pair, each candidate property is scored by translation distance: $\text{score}(h, r, t) = 1 / (1 + \| \mathbf{h} + \mathbf{r} - \mathbf{t} \|)$, evaluated in both directions. Variants include random versus sentence-transformer embedding initialization and whether hierarchy triples are included in training. While TransE is a classic \emph{structure-based} knowledge graph completion method, it still has the capacity of including surface names through initialization with sentence embedding. However, the method is \textit{not adapted for context enrichment}.

\para{Pretrained MLM prompting.} For each class pair $(c_i, c_j)$, the framework constructs sentences in both directions (``$c_i$ [MASK] $c_j$'' and ``$c_j$ [MASK] $c_i$'') and scores each candidate relation by its probability at the masked position using \texttt{distilbert-base-uncased} pretrained MLM.
Multi-token property names are handled by averaging subword probabilities.
This heuristic we develop supports context enrichment: metric descriptions or comments can be prepended to the template.

\subsubsection{Task 3 --- Data Property Assignment}
We re-use Task~2 models (TransE and BertConvE) by augmenting their training triples with data-property axioms encoded as $(c, a, \tau)$, where each XSD datatype $\tau$ (e.g.\ \texttt{xsd:integer}) becomes a synthetic tail entity shared across all properties of that datatype. For BertConvE the xsd-type tails are marked walk-terminating so they do not act as transit hubs in the random walks. At inference, the head slot is masked (``[MASK] $a$ $\tau$'') and candidate classes are ranked by translation distance (TransE) or masked-token probability at the head position (BertConvE) -- \textit{no context enrichment is used for TransE or BertConvE}. 

\para{Pretrained MLM prompting.} A [MASK] is placed in a template ``[MASK] can have \{property\}'' by default.
Metric context that provides target values is used to modify the template as ``[MASK] is \{property\}'' for booleans, or ``The \{property\} of [MASK] is \{targetValue\}''.
Candidates are scored by the average MLM probability of their subword tokens at the mask using \texttt{distilbert-base-uncased} pretrained model.
Metric descriptions or comments can be prepended to the template for context enrichment.

\begin{table*}[ht]
\small
\centering
\caption{RQ1 --- Relation type prediction: algorithm comparison across ontologies. Mean$\pm$std over 5 random seeds. Best score per ontology is in boldface, and statistically significant wins are underlined.}
\label{tab:rq1-objprop}
\begin{tabular}{@{}lllllll@{}}
\toprule
\textbf{Ontology} & \textbf{Algorithm}            & \textbf{MRR}                & \textbf{Hits@1}             & \textbf{Hits@3}             & \textbf{Hits@5}             & \textbf{Hits@10}            \\ \midrule
Pizza             & BertConvE                       & \underline{$\mathbf{0.73_{\pm 0.09}}$}  & $\mathbf{0.49_{\pm 0.18}}$  & $\mathbf{0.98_{\pm 0.02}}$  & $\mathbf{0.99_{\pm 0.01}}$  & $\mathbf{1.00_{\pm 0.00}}$  \\
Pizza             & TransE (Random Init)           & $0.47_{\pm 0.11}$           & $0.23_{\pm 0.13}$           & $0.63_{\pm 0.19}$           & $0.95_{\pm 0.03}$           & $\mathbf{1.00_{\pm 0.00}}$  \\
Pizza             & TransE (ST init)               & $0.35_{\pm 0.11}$           & $0.10_{\pm 0.11}$           & $0.40_{\pm 0.17}$           & $0.76_{\pm 0.21}$           & $\mathbf{1.00_{\pm 0.00}}$  \\
Pizza             & MLM                            & $0.45_{\pm 0.03}$           & $0.19_{\pm 0.04}$           & $0.66_{\pm 0.11}$           & $0.94_{\pm 0.02}$           & $\mathbf{1.00_{\pm 0.00}}$  \\\midrule
FIBO$_{BE}$          & BertConvE                       & $0.24_{\pm 0.13}$           & $0.12_{\pm 0.16}$           & $\mathbf{0.28_{\pm 0.20}}$  & $\mathbf{0.40_{\pm 0.18}}$  & $\mathbf{0.48_{\pm 0.16}}$  \\
FIBO$_{BE}$          & TransE (Random Init)           & $0.15_{\pm 0.09}$           & $0.04_{\pm 0.08}$           & $0.12_{\pm 0.16}$           & $0.24_{\pm 0.20}$           & $0.40_{\pm 0.25}$           \\
FIBO$_{BE}$          & TransE (ST init)               & $\mathbf{0.25_{\pm 0.09}}$  & $\mathbf{0.16_{\pm 0.08}}$  & $0.24_{\pm 0.15}$           & $0.28_{\pm 0.16}$           & $0.36_{\pm 0.15}$           \\
FIBO$_{BE}$          & MLM                            & $0.21_{\pm 0.10}$           & $0.12_{\pm 0.10}$           & $0.12_{\pm 0.10}$           & $0.24_{\pm 0.15}$           & $\mathbf{0.48_{\pm 0.10}}$  \\\midrule
SAREF$_{ener}$       & BertConvE                       & $0.14_{\pm 0.04}$           & $0.07_{\pm 0.04}$           & $0.18_{\pm 0.07}$           & $0.23_{\pm 0.08}$           & $0.30_{\pm 0.06}$           \\
SAREF$_{ener}$       & TransE (Random Init)           & $0.02_{\pm 0.01}$           & $0.00_{\pm 0.00}$           & $0.02_{\pm 0.02}$           & $0.02_{\pm 0.02}$           & $0.02_{\pm 0.02}$           \\
SAREF$_{ener}$       & TransE (ST init)               & $\mathbf{0.31_{\pm 0.04}}$  & $\mathbf{0.21_{\pm 0.06}}$  & $\mathbf{0.36_{\pm 0.05}}$  & $\mathbf{0.43_{\pm 0.07}}$  & $\mathbf{0.51_{\pm 0.07}}$  \\
SAREF$_{ener}$       & MLM                            & $0.12_{\pm 0.02}$           & $0.05_{\pm 0.03}$           & $0.10_{\pm 0.03}$           & $0.21_{\pm 0.05}$           & $0.30_{\pm 0.04}$           \\\midrule
JRC        & BertConvE                       & $0.26_{\pm 0.03}$           & $0.04_{\pm 0.04}$           & $0.25_{\pm 0.05}$           & $0.52_{\pm 0.06}$           & $\mathbf{0.96_{\pm 0.01}}$  \\
JRC        & TransE (Random Init)           & $0.16_{\pm 0.02}$           & $0.02_{\pm 0.01}$           & $0.11_{\pm 0.03}$           & $0.21_{\pm 0.05}$           & $0.55_{\pm 0.07}$           \\
JRC        & TransE (ST init)               & $0.16_{\pm 0.06}$           & $0.04_{\pm 0.04}$           & $0.12_{\pm 0.10}$           & $0.21_{\pm 0.12}$           & $0.49_{\pm 0.15}$           \\
JRC        & MLM                            & \underline{$\mathbf{0.38_{\pm 0.02}}$}  & $\mathbf{0.22_{\pm 0.02}}$  & $\mathbf{0.44_{\pm 0.02}}$  & $\mathbf{0.57_{\pm 0.02}}$  & $0.69_{\pm 0.02}$           \\\midrule
CSO$_{sec}$     & BertConvE                       & \underline{$\mathbf{0.72_{\pm 0.01}}$}  & $\mathbf{0.46_{\pm 0.02}}$  & $\mathbf{0.97_{\pm 0.01}}$  & $\mathbf{1.00_{\pm 0.00}}$  & $\mathbf{1.00_{\pm 0.00}}$  \\
CSO$_{sec}$     & TransE (Random Init)           & $0.54_{\pm 0.02}$           & $0.29_{\pm 0.03}$           & $0.74_{\pm 0.05}$           & $\mathbf{1.00_{\pm 0.00}}$  & $\mathbf{1.00_{\pm 0.00}}$  \\
CSO$_{sec}$     & TransE (ST init)               & $0.51_{\pm 0.02}$           & $0.23_{\pm 0.01}$           & $0.75_{\pm 0.08}$           & $\mathbf{1.00_{\pm 0.00}}$  & $\mathbf{1.00_{\pm 0.00}}$  \\
CSO$_{sec}$     & MLM                            & $0.52_{\pm 0.01}$           & $0.25_{\pm 0.02}$           & $0.75_{\pm 0.02}$           & $\mathbf{1.00_{\pm 0.00}}$  & $\mathbf{1.00_{\pm 0.00}}$  \\\midrule
TAC               & BertConvE                       & $0.22_{\pm 0.05}$           & $0.18_{\pm 0.05}$           & $0.21_{\pm 0.06}$           & $0.24_{\pm 0.05}$           & $0.33_{\pm 0.08}$           \\
TAC               & TransE (Random Init)           & $0.02_{\pm 0.01}$           & $0.00_{\pm 0.00}$           & $0.00_{\pm 0.00}$           & $0.02_{\pm 0.03}$           & $0.03_{\pm 0.03}$           \\
TAC               & TransE (ST init)               & $\mathbf{0.29_{\pm 0.07}}$  & $\mathbf{0.22_{\pm 0.08}}$  & $\mathbf{0.33_{\pm 0.08}}$  & $\mathbf{0.36_{\pm 0.05}}$  & $\mathbf{0.38_{\pm 0.07}}$  \\
TAC               & MLM                            & $0.20_{\pm 0.05}$           & $0.12_{\pm 0.05}$           & $0.18_{\pm 0.05}$           & $0.27_{\pm 0.07}$           & $0.37_{\pm 0.06}$           \\\midrule
CertGraph         & BertConvE                       & \underline{$\mathbf{0.38_{\pm 0.05}}$}  & $\mathbf{0.17_{\pm 0.03}}$  & $\mathbf{0.48_{\pm 0.07}}$  & $\mathbf{0.65_{\pm 0.12}}$  & $\mathbf{0.83_{\pm 0.06}}$  \\
CertGraph         & TransE (Random Init)           & $0.12_{\pm 0.01}$           & $0.01_{\pm 0.01}$           & $0.07_{\pm 0.04}$           & $0.26_{\pm 0.08}$           & $0.32_{\pm 0.07}$           \\
CertGraph         & TransE (ST init)               & $0.13_{\pm 0.06}$           & $0.02_{\pm 0.03}$           & $0.13_{\pm 0.08}$           & $0.19_{\pm 0.12}$           & $0.35_{\pm 0.17}$           \\
CertGraph         & MLM                            & $0.19_{\pm 0.04}$           & $0.10_{\pm 0.04}$           & $0.15_{\pm 0.05}$           & $0.18_{\pm 0.05}$           & $0.37_{\pm 0.10}$           \\ \bottomrule
\end{tabular}%

\end{table*}

\subsection{Results}
\label{sec:results}

\subsubsection{RQ1 --- Algorithm Comparison}
\label{sec:rq1}
\textbf{Task~1}
results are presented in Table~\ref{tab:rq1-parent} on all datasets.
Applying the paired significance test, only two results survive: ChildAgg outranks TaxoExpan on JRC, and TaxoExpan outranks ChilAggS on CSO$_{sec}$.
The remaining ontologies are statistical ties: Pizza, CertGraph and TAC, while the FIBO$_{BE}$ and SAREF$_{ener}$ flip sign across training seeds and therefore fall within TaxoExpan's seed variance.
The two robust outcomes track the structural signal available for TaxoExpan's GNN.
CSO$_{sec}$ is by far the largest ontology, where its learned ego-graph matching has both signal and candidate volume, whereas JRC, with only two-level hierarchy depth, does not offer meaningful ego-graphs, making ChildAgg's frozen embeddings hard to beat.

\begin{table}[]
\small
\centering
\caption{RQ1 --- Data property assignment: algorithm comparison across ontologies. Mean$\pm$std over 5 random seeds. Ontologies with no attributes are excluded. Best score per ontology is in boldface, and statistically significant wins are underlined.}
\label{tab:rq1-dataprop}
\begin{tabular}{@{}lll@{}} 
\toprule
\textbf{Ontology} & \textbf{Algorithm}            & \textbf{Accuracy}             \\ \midrule
SAREF$_{ener}$  & BertConvE                       & $0.56_{\pm 0.03}$             \\
SAREF$_{ener}$  & TransE (Random Init)           & $0.50_{\pm 0.05}$             \\
SAREF$_{ener}$  & TransE (ST init)               & $0.47_{\pm 0.05}$             \\
SAREF$_{ener}$  & MLM                            & $\mathbf{0.64_{\pm 0.06}}$    \\\midrule
JRC    & BertConvE                       & \underline{$\mathbf{0.99_{\pm 0.00}}$}    \\
JRC    & TransE (Random Init)           & $0.52_{\pm 0.10}$             \\
JRC    & TransE (ST init)               & $0.96_{\pm 0.01}$             \\
JRC    & MLM                            & $0.52_{\pm 0.01}$             \\\midrule
TAC          & BertConvE                       & $0.51_{\pm 0.02}$             \\
TAC          & TransE (Random Init)           & $0.54_{\pm 0.02}$             \\
TAC          & TransE (ST init)               & $0.46_{\pm 0.03}$             \\
TAC          & MLM                            & \underline{$\mathbf{0.62_{\pm 0.02}}$}    \\\midrule
CertGraph    & BertConvE                       & $\mathbf{0.59_{\pm 0.03}}$    \\
CertGraph    & TransE (Random Init)           & $0.53_{\pm 0.06}$             \\
CertGraph    & TransE (ST init)               & $0.50_{\pm 0.03}$             \\
CertGraph    & MLM                            & $0.51_{\pm 0.02}$             \\ \bottomrule
\end{tabular}
\end{table}

\begin{table*}[ht]
\small
\centering
\caption{RQ2 --- Context enrichment of ChildAgg for parent class prediction (Task~1) on CertGraph.}
\label{tab:rq2-parent}
\begin{tabular}{@{}lrrrrr@{}}
\toprule
\textbf{Context} & \textbf{MRR} & \textbf{Hits@1} & \textbf{Hits@3} & \textbf{Hits@5} & \textbf{Hits@10} \\
\midrule
No context                          & \textbf{0.49} & \textbf{0.44} & \textbf{0.51} & \textbf{0.58} & \textbf{0.60} \\
\midrule
+ Ontology comments                 & 0.43 & 0.37 & 0.48 & 0.54 & \textbf{0.60} \\
\midrule
+ Metric description                & 0.34 & 0.25 & 0.38 & 0.48 & 0.53 \\
+ Metric comment                    & 0.29 & 0.16 & 0.42 & 0.46 & 0.52 \\
+ Metric description + comment      & 0.32 & 0.20 & 0.42 & 0.49 & 0.53 \\
\midrule
All context (metric + ontology)     & 0.34 & 0.23 & 0.44 & 0.48 & 0.52 \\
\bottomrule
\end{tabular}
\end{table*}

\begin{table*}[!t]
\small
\centering
\caption{RQ2 --- Context enrichment of MLM prompting for relation type prediction: (Task~2) on CertGraph.}
\label{tab:rq2-objprop}
\begin{tabular}{@{}lrrrrr@{}}
\toprule
\textbf{Context} & \textbf{MRR} & \textbf{Hits@1} & \textbf{Hits@3} & \textbf{Hits@5} & \textbf{Hits@10} \\
\midrule
No context                          & 0.21 & 0.10 & 0.20 & 0.30 & 0.40 \\
\midrule
+ Ontology comments                 & 0.21 & 0.10 & 0.20 & \textbf{0.40} & \textbf{0.50} \\
\midrule
+ Metric description                & \textbf{0.27} & \textbf{0.20} & \textbf{0.30} & 0.30 & 0.40 \\
+ Metric comment                    & 0.21 & 0.10 & 0.20 & 0.30 & 0.30 \\
+ Metric description + comment      & 0.25 & \textbf{0.20} & 0.20 & 0.30 & 0.30 \\
\midrule
All context (metric + ontology)     & 0.22 & 0.10 & \textbf{0.30} & 0.30 & 0.40 \\
\bottomrule
\end{tabular}
\end{table*}

\begin{table}[!t]
\small
\centering
\caption{RQ2 --- Context enrichment of MLM prompting for data property assignment (Task~3) on CertGraph.}
\label{tab:rq2-dataprop}
\begin{tabular}{@{}lr@{}}
\toprule
\textbf{Context} & \textbf{Accuracy} \\
\midrule
No context                          & 0.71 \\
\midrule
+ Ontology comments                 & 0.64 \\
\midrule
+ Metric description                & 0.68 \\
+ Metric comment                    & \textbf{0.82} \\
+ Metric description + comment      & 0.79 \\
\midrule
All context (metric + ontology)     & 0.68 \\
\bottomrule
\end{tabular}
\end{table}

\textbf{Task~2} results shown in Table~\ref{tab:rq1-objprop} show a significant outperformance of BertConvE in terms of MRR on Pizza, CSO$_{sec}$, and CertGraph.
Templated MLM wins on JRC, where the (domain, relation, range) triple reads as a natural English clause.
TransE (ST init) results on FIBO$_{BE}$, SAREF$_{ener}$, and TAC are statistical ties.
The randomly-initialized TransE is rarely competitive, suggesting that surface lexical information are more useful than pure structural training at these graph sizes.
CSO$_{sec}$ saturates Hits@5 and Hits@10 and Pizza saturates Hits@10 due to the low number of relation types: two types for CSO and three for Pizza, both bidirectional.

\textbf{Task~3} results, shown in Table~\ref{tab:rq1-dataprop}, show two significant results: BertConvE outperform the rest on JRC and MLM propmting outperforms the rest on TAC.
BertConvE on CertGraph falls within the training noise.
The JRC result stands out, as BertConvE reaches $0.99$ accuracy while both TransE variants and MLM hover around the $0.50$ random baseline on the same ontology.
Since Pizza and CSO$_{sec}$ have no attributes, and FIBO$_{BE}$ only has three classes with attributes, they are excluded from this experiment.

\subsubsection{RQ2 --- Context Enrichment}
\label{sec:rq2}

To answer the context enrichment question, we apply algorithms, which are capable of incorporating external context into their predictions, that is ChildAgg for Task~1 (Table~\ref{tab:rq2-parent}) and MLM prompting for both Task~2 (Table~\ref{tab:rq2-objprop}) and Task~3 (Table~\ref{tab:rq2-dataprop}).
As all algorithms are in inference mode and do not require training, there is no setup for multiple random seeds.
To focus on the contribution of ontology and metric context, we limit the test set in this experiment to the classes and properties mentioned in the metric catalogue associated with CertGraph -- resulting in 81 cases for Task~1, 10 cases for Task~2, and 28 cases for Task~3.
For each task, we evaluate the respective method with no context, with ontology context represented by \texttt{rdfs:comment} elements, and with metric-driven context represented by both the description and comment in the metric file.

\textbf{Task~1:} Table~\ref{tab:rq2-parent} shows that additional context does not increase the score of ChildAgg on Task~1, as no context achieves an MRR of $0.49$. 
However, the MRR drop with the ontology-context ($-6$~pp) is smaller than that with the metric context ($-15..-20$~pp).

\textbf{Task~2:} Table~\ref{tab:rq2-objprop} shows that the metric-driven context increases the score significantly for Tasks~2 with an MRR gain of $6$~pp, though the ontology-driven context achieves better Hits@5 and Hits@10.

\textbf{Task~3:} In Table~\ref{tab:rq2-dataprop}, the same pattern repeats with the metric comment alone lifting the accuracy from $0.71$ to $0.82$ ($+11$~pp).
Ontology comments appear to have and a mildly harmful effect ($-7$~pp) for Task~3.
We discuss these results and their implications in the next section.

%
\section{Discussion}
\label{sec:discussion}
\subsection{Implications}
\para{Structured metric catalogues are an underexploited input modality.}
Prior ontology-extension work mostly draws on ontology structure~\cite{shen_taxoexpan_2020} and external natural-language documentation~\cite{sanagavarapu2022ontoenricher}. Structured metric definitions provide an independent text source that boosts property and relation prediction scores on CertGraph in comparison to bare class names.
The RQ2 results shows a $6$~pp MRR improvement on Task~2 ($0.21 \to 0.27$) and a $+11$~pp Task~3 accuracy lift ($0.71 \to 0.82$) once metric description and comment are added.
These results highlight the benefit of this input modality as a context for language-based models for ontology extension.

\para{More context is not universally beneficial.}
The same metric prose that helps Task~2 and Task~3 \emph{hurts} parent class prediction in Task~1 (MRR $0.49 \to 0.34$, $\Delta = -0.15$).
We hypothesize that the quality of the added context plays a major role here.
In particular, the text to be expected in metrics associated with CertGraph usually describes the relation between two or more classes and how their properties should be configured.
This context aligns with the MLM prompt for Tasks~2 and~3.
However, this description does not align with Task~1, where a class-level context could be more beneficial.
In a verification experiment, we measure the cosine similarity between the query class ST-embedding and its aggregate siblings embedding, which averages at 0.533 with no context.
Appending context shifts the similarity down to 0.287 for metric description, to 0.383 for metric comments and to 0.524 for ontology comments.
The farther a query's embedding from its ground-truth aggregate embedding, the less likely the true parent will be recommended.
This aligns with the MRR reduction caused by the context on Task~1 (Table~\ref{tab:rq2-parent}).

\para{Practical low-friction industry adoption.}
organizations already maintain operational metric catalogues in structured form (YAML, JSON, RDF).
COntExt makes these immediately reusable as ontology-extension inputs with no additional annotation effort.
This is particularly beneficial in the cybersecurity field, where adopting security metrics has been receiving growing attention.
The way the context is extracted and integrated can be easily configured within this framework.

\subsection{Limitations}
\para{Single ontology for RQ2.} The metric-context findings rest on a single (ontology, metric corpus) pair, CertGraph and its cybersecurity-certification metric catalogue, because no other ontology in our evaluation suite is shipped with a paired metric corpus of comparable structure. Replicating the metric-enrichment pattern on another domain pair is necessary to argue that the effect generalizes beyond CertGraph's particular YAML schema and prose style.

\para{Metric context not extended to structural methods.}
The RQ2 comparison evaluates metric description and comment with text-template scorers only, MLM and ChildAgg. The structural methods, TransE, BertConvE, are absent because they consume tokenized triples rather than free-form prose and do not natively expose a slot for auxiliary text.
Adapting such structural methods to ingest metric content would require separate research efforts.

\para{No large-language-model baseline.}
As our work aims to exploit operational metrics efficiently, we do not include LLMs in our algorithm choices despite their known effectiveness.
While some previous work suggests that LLMs still lag behind for ontological consistency and they require more validation \cite{kollapally2025ontology,dong_language_2024}, evaluating LLMs with metrics-driven context would still be worth investigating.

\section{Conclusion}
\label{sec:conclusion}

In this paper, we have introduced an algorithm-agnostic context-aware framework, COntExt, which leverages operational metrics and other input modalities as context for the ontology extension task.
Our evaluation shows that operational metrics provide viable context to the ontology extension task, demonstrating a significant improvement over ontology-driven context for relation type prediction and data property assignment tasks.
Organizations can use this framework to support their ontology maintenance, with the ability to harness relevant context, such as metrics catalogues and KPIs.
Future work will develop tailored algorithms for metrics-oriented ontology extension as well as a comprehensive user study involving domain experts to evaluate the efficiency and utility of COntExt framework for semi-automated ontology extension.

\section*{Acknowledgments}
This work was funded by the Horizon Europe project
EMERALD, grant agreement ID 101120688.
We thank Mark Kröll for insightful
discussions and valuable feedback on this work.

\para{Generative AI use.}
We have used Anthropic's Claude Code \cite{anthropic2026claudeopus47} to write the python scripts according to our experimental design.
All the AI-generated code was thoroughly reviewed by the corresponding author at every step.
We have also used Claude Code to rewrite parts of the introduction (Section~\ref{sec:introduction}) and to
formalize the descriptions of the framework (Section~\ref{sec:framework}) and the evaluation setup (Sections~\ref{sec:protocols}--\ref{sec:algorithms}).
AI-generated text was reviewed and often significantly changed by the authors.
\bibliographystyle{apalike}
{\small
\bibliography{references}}

\end{document}